\documentclass[lettersize,journal]{IEEEtran}
\usepackage{amsmath,amsfonts}
\usepackage{algorithmic}
\usepackage{algorithm}
\usepackage{array}
\usepackage[caption=false,font=normalsize,labelfont=sf,textfont=sf]{subfig}
\usepackage{textcomp}
\usepackage{stfloats}
\usepackage{url}
\usepackage{verbatim}
\usepackage{graphicx}
\usepackage{cite}
\usepackage{xcolor}
\usepackage{CJKutf8}
\usepackage[T1]{fontenc}
\usepackage{xcolor}
\usepackage{makecell} % 引入makecell包
\usepackage{multirow}  % 用于表格的纵向合并行
\usepackage{pgfplots}
\usepackage{caption}   % 支持 \captionof
\usepackage{booktabs} % 用于优化表格线条

\begin{document}
\begin{CJK}{UTF8}{gbsn}
\title{KB-DMGen: Knowlege-Based Global Guidance and Dynamic Pose Masking for Human Image Generation}

\author{Shibang Liu, Xuemei Xie,~\IEEEmembership{Senior Member,~IEEE} and Guangming Shi,~\IEEEmembership{Fellow,~IEEE}}
%\IEEEpubid{\begin{minipage}{\textwidth}\ \centering 
%		This work has been submitted to the IEEE for possible publication. \\
%	Copyright may be transferred without notice, after which this version may no longer be accessible
%\end{minipage}}

%\maketitle

\twocolumn[{%
	\renewcommand\twocolumn[1][]{#1}%
	\maketitle
	\begin{center}
		    \centering
		\includegraphics[width=0.8\textwidth]{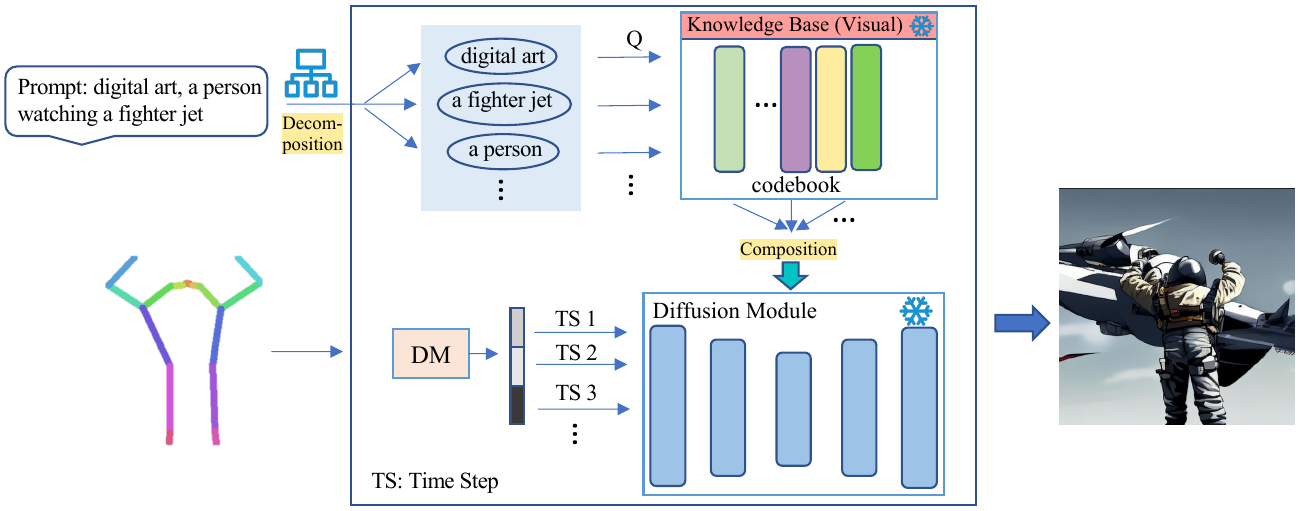}
		    \captionof{figure}{Overview of our method. The visual Knowledge Base (KB) provides global guidance by encoding visual features related to the text description, while the Dynamic pose Mask (DM) enables fine-grained local control. Together, they achieve unified control of both global semantics and local details.}
		    \label{fig1}
		\end{center}%
	}]

\begin{abstract}
Recent methods using diffusion models have made significant progress in Human Image Generation (HIG) with various control signals such as pose priors. In HIG, both accurate human poses and coherent visual quality are crucial for image generation. However, most existing methods mainly focus on pose accuracy while neglecting overall image quality, often improving pose alignment at the cost of image quality. To address this, we propose Knowledge-Based Global Guidance and Dynamic pose Masking for human image Generation (KB-DMGen). The Knowledge Base (KB), implemented as a visual codebook,  provides coarse, global guidance based on input text-related visual features, improving pose accuracy while maintaining image quality, while the Dynamic pose Mask (DM) offers fine-grained local control to enhance precise pose accuracy. By injecting KB and DM at different stages of the diffusion process, our framework enhances pose accuracy through both global and local control without compromising image quality. Experiments demonstrate the effectiveness of KB-DMGen, achieving new state-of-the-art results in terms of AP and CAP on the HumanArt dataset.  The project page and code are available at \url{https://lushbng.github.io/KBDMGen}.

\end{abstract}


\section{Introduction}
\label{sec:intro}

\IEEEPARstart{T}{he} goal of human image generation (HIG) is to synthesize high-quality images under certain conditions based on a series of prompts (e.g., pose)~\cite{Humansd,StablePose,ControlNet,T2iAdapte}. Human image generation serves a wide range of real-world applications, including animation~\cite{Vlogger}, game production~\cite{SCStory}, and other fields.

Previous methods~\cite{ControllableGan,PGPIG,Xinggan} require a source image during training using variational autoencoders (VAEs)~\cite{VAE} or Generative Adversarial Networks (GANs)~\cite{GAN} for dictating the style of the generated images. These methods synthesize target images with specific human features by adjusting the source images, but the training process of these methods is unstable and highly dependent on the distribution of the source images. In contrast, recent advances in controllable text-to-image (T2I) Stable Diffusion (SD)~\cite{StableDiffusion} show potential to eliminate the need for source images, enabling greater creative freedom through reliance on text prompts and external conditions~\cite{ControlNet,UniControlnet,T2iAdapte,Gligen}. These methods often face challenges in accurately matching conditional images with sparse representations such as skeleton pose data~\cite{Humansd}. To achieve accurate pose control, various pose guided T2I methods are proposed, such as introducing pose heatmap supervision loss~\cite{Humansd},  establishing a graph
topological structure between the pose priors and latent representation of diffusion models~\cite{Grpose}, applying pose masks to the attention module of the ViT~\cite{StablePose}. These strategies effectively guide the network to focus on pose regions, thereby improving pose fidelity.\IEEEpubidadjcol

While precise pose alignment is essential, high-quality human image generation also demands the guidence of global visual semantics to ensure overall image quality. However, these methods~\cite{Grpose,Humansd,StablePose} emphasize the modeling of pose details while neglecting overall image quality. To address this issue, we propose Knowledge Based Global Guidance and Dynamic pose Masking for Human Image Generation (KB-DMGen). KB-DMGen introduces a visual Knowledge Base (KB) to provide global visual semantics related to the input text during image generation, improving pose accuracy while maintaining image quality. Meanwhile, the Dynamic Masking (DM) mechanism adaptively adjusts the weights of pose-related regions, enabling the model to better balance local pose precision compared with Stable-pose~\cite{StablePose}.  By injecting them into different stages of the diffusion process, our framework enhances pose accuracy through both global and local aspects without compromising image quality. In summary, the contributions of this paper are as follows:
\begin{itemize}
	\item Designing a visual KB to improve pose generation accuracy while preserving image quality.
	\item Designing DM to control the precise generation of poses.
    \item Injecting KB and DM into different stages of the SD~\cite{StableDiffusion} to enhance pose accuracy via
    global  and local control without compromising image quality.

\end{itemize}

\begin{figure}[t]
	\centering
	
	\includegraphics[width=1\linewidth]{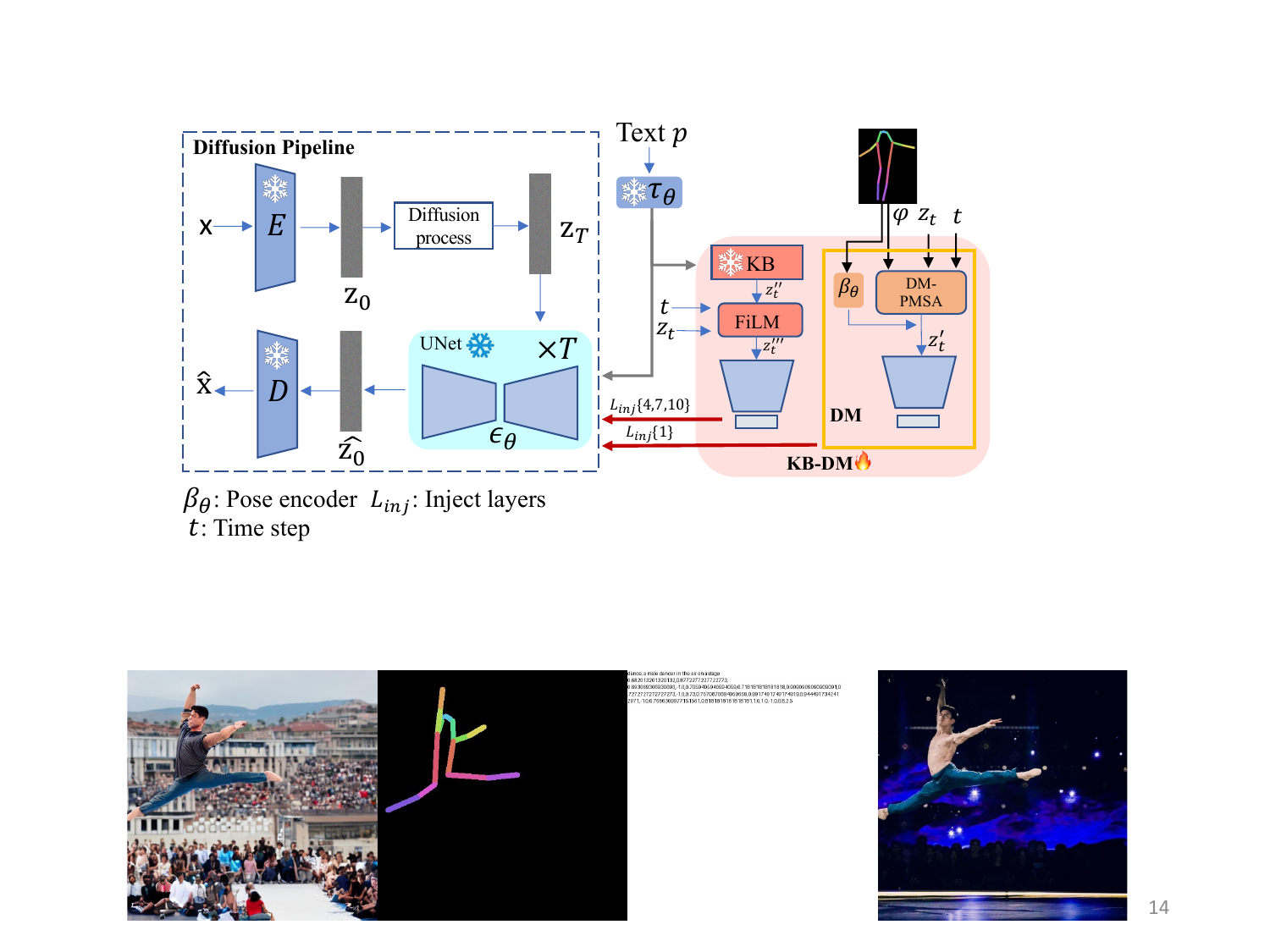}
	
	\caption{The architecture of  KB-DMGen. The text encoding is used to retrieve semantic codebook features from a visual Knowledge Base (KB), providing global semantic guidance; meanwhile, pose information generates temporally dynamic masks through the diffusion process, enabling precise control over human pose.}
	\label{fig2}
\end{figure}

\begin{figure}[!t]
	\centering
	\subfloat[]{\includegraphics[width=3.2in]{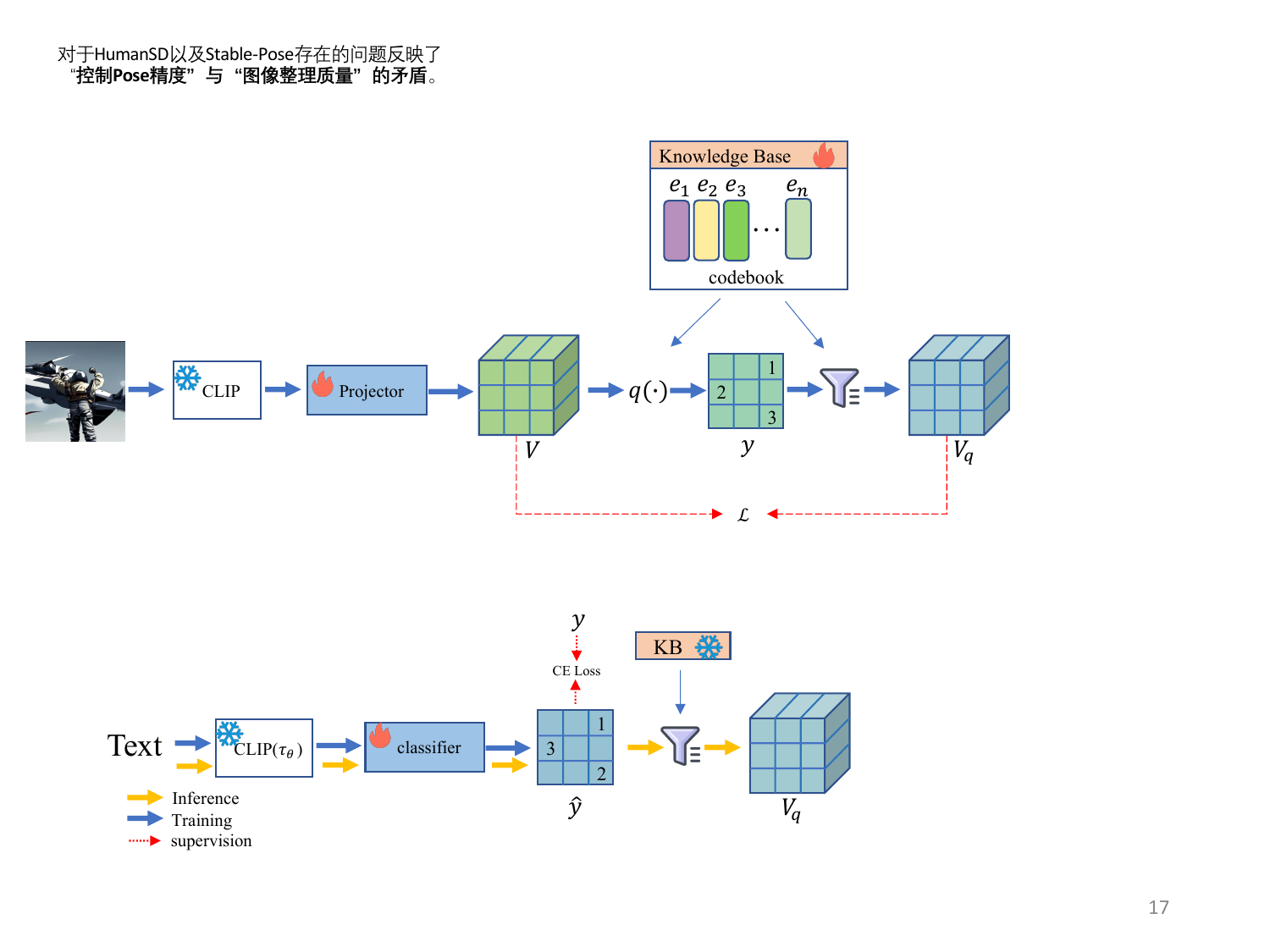}%
		\label{fig3-a}}
	\hspace{0.6cm}
	\subfloat[]{\includegraphics[width=3.1in]{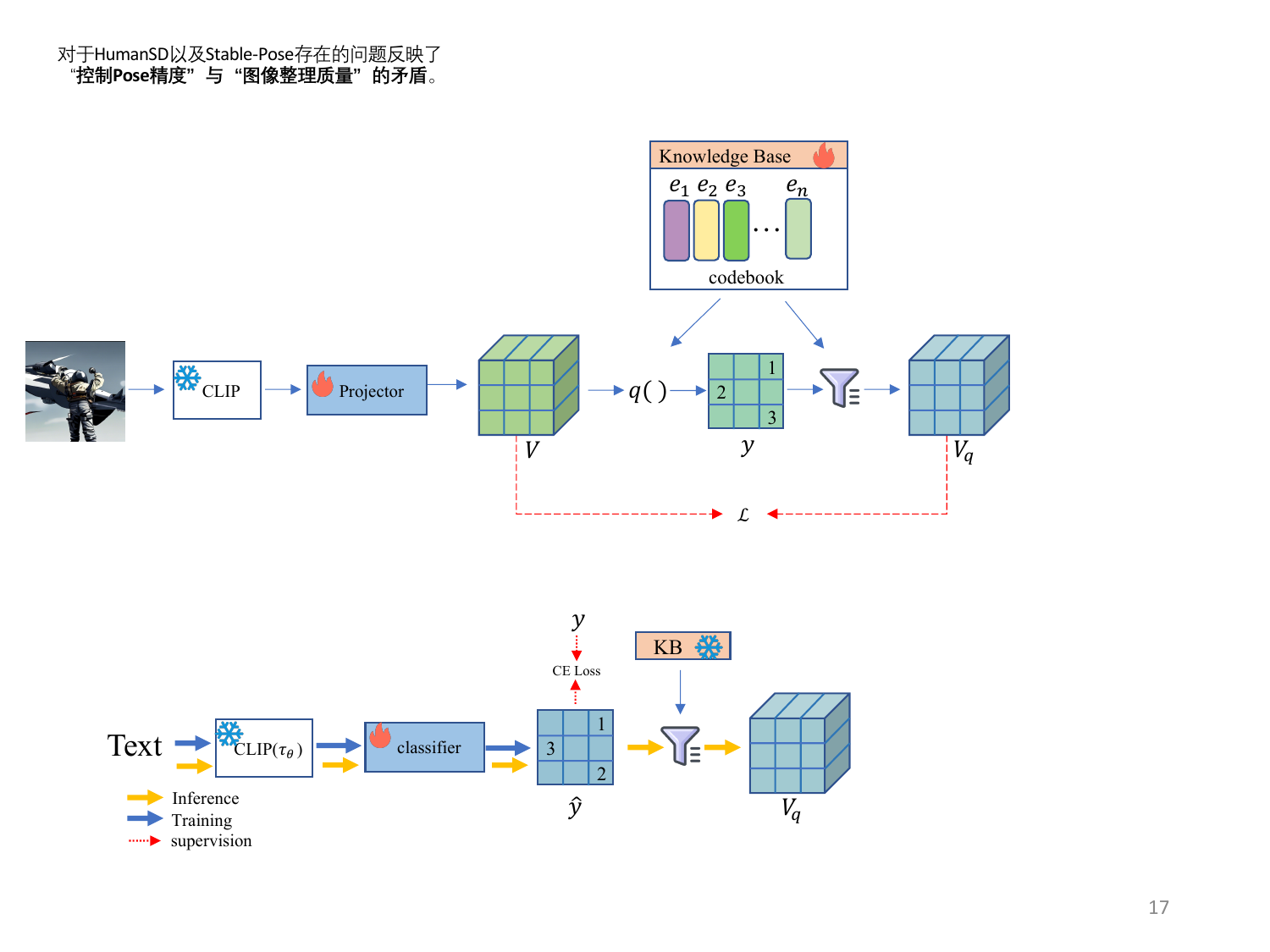}%
		\label{fig3-b}}
	\caption{(a) The codebook training. $V$ is the image feature after mapping. $q(\cdot)$ denotes the nearest-neighbor mapping of $V$ in the codebook embedding space, where $y$ is the nearest neighbor index and $V_q$ is the image feature quantized by the codebook. (b) The process of classifier training. The text encoder is consistent with $\tau_\theta$ in Fig.~\ref{fig2}. $\hat{y}$ denotes the codebook entry assignment of visual tokens predicted from text features, supervised by $y$ obtained in (a).}
	\label{fig3}
\end{figure}

\section{Related work}
\label{sec:rel}
{\bf{Pose-Guided Human Iage Generation.}} Previous pose-guided approaches take a source image and pose as input to generate photo-realistic human images with preserved appearance, relying on GANs/VAEs for conditional generation~\cite{ControllableGan,Xinggan,PGPIG,MustGan,HPTRN}. Recently, Bhunia et al.~\cite{PIDM} introduce texture diffusion modules and disentangled classifier-free guidance to accurately model appearance-pose relationships and ensure input-output consistency. Zhang et al.~\cite{DualTaskCorrelation} propose a novel Dual-task Pose Transformer Network (DPTN), which introduces an auxiliary task (i.e., source to source task) and exploits the dual-task correlation to promote the performance of PGPIG. Shen et al.~\cite{AdvancingpProgressiveConditional} presented a Progressive Conditional Diffusion Model (PCDM), which narrows the gap between character images under target poses and source poses step by step via a three-stage procedure. The above methods~\cite{PIDM,DualTaskCorrelation,AdvancingpProgressiveConditional} remain highly dependent on the distribution of source images, operating under the paradigm of pose transfer that necessitates original image inputs. In contrast, recent work based on pose guided T2I dispenses with the need for source images entirely: HumanSD~\cite{Humansd} enhances the pose accuracy of HIG via heatmap-guided losses and Stable-Pose~\cite{StablePose} employs coarse-to-fine masking for precise HIG. GRPose~\cite{Grpose} establish a graph topological structure between the pose priors and latent representation of diffusion models to capture the intrinsic associations between different pose parts. However, these methods~\cite{Humansd,StablePose,Grpose} prioritize pose fidelity and lack global image quality assurance, prompting us to introduce a visual knowledge base, which enhances the pose accuracy of HIG and ensures image quality via global guidance and a dynamic mask to guarantee pose precision via local control.

{\bf{Controllable Diffusion Models.}} Large-scale T2I diffusion models~\cite{StableDiffusion,ZeroShotT2I,HTClipLatents,PhotorealisticT2I,Glide} excel in generating diverse high-quality images but lack precision with text-only prompts. Recent studies have enhanced their controllability using conditions such as edges, sketches, and human poses~\cite{Composer,ControlNet,T2iAdapte,Gligen,UniControlnet,Humansd,Controlnet++,Controlling}, primarily through two approaches: training full T2I models (e.g., Composer~\cite{Composer} constructing diffusion models via decomposition-composition for multi-control, HumanSD~\cite{Humansd} fine-tuning SD with pose-specific losses) or developing plug-in adapters for pre-trained models (e.g., T2I-Adapter~\cite{T2iAdapte} and GLIGEN~\cite{Gligen} integrating lightweight adapters into frozen SD, ControlNet~\cite{ControlNet} encoding conditions via trainable SD encoder copies, Uni-ControlNet~\cite{UniControlnet} enabling multi-scale conditional injection, and ControlNet++~\cite{Controlnet++} optimizing cycle consistency). Our approach aligns with the adapter-based paradigm, freezing the pre-trained SD model.

{\bf{Relation to VQ-VAE~\cite{NDRL}.}} The core idea of Vector Quantised-Variational AutoEncoder (VQ-VAE)\cite{NDRL,TamingIS,AutoencodingSimilarity} is to learn discrete latent representations by mapping continuous features into a codebook of quantized vectors. This foundational framework has been used in various tasks. For example, PCTPose\cite{PCTPose} uses a discrete codebook to model 2D human joint relations, while VQ-VAE-2~\cite{VQVAE2} extends VQ-VAE for large-scale image generation. These methods~\cite{VQVAE2,PCTPose} employ EMA updates to mitigate codebook collapse. However, such updates rely on additional hyperparameters (e.g., decay rates) and adapt slowly. Moreover, the pixel-wise sampling in VQVAE2~\cite{VQVAE2} is time-consuming. In addition, Zero-shot T2I~\cite{ZeroShotT2I} explores VQ-VAE-based frameworks for text-to-image generation, and Text2Human~\cite{Text2human} uses a VQ-VAE-style codebook to model local clothing attributes in human image generation. In contrast, during the separate training of KB, our method focuses on aligning language and vision through a shared discrete token space. Specifically, we discretize image features to form a visual semantic codebook, introduce an entropy-based loss to prevent codebook collapse, and train a classifier to retrieve relevant tokens from the visual semantic codebook based on textual prompts, providing more efficient semantic guidance for image generation.

\section{Method}
\label{sec:meth}
This section presents the overall framework, followed by detailed codebook in the KB and classifier, concludes with the DM strategy, and finally introduces the loss of KB-DMGen.

\subsection{Overview of KB-DMGen}
Our goal is to generate high-quality human images conditioned on pose priors. To this end, we propose KB-DMGen, a Stable Diffusion based framework equipped with two adapters: a KB Adapter and a DM Adapter (Fig.~\ref{fig2}). These modules enhance pose accuracy while preserving global image fidelity. Specifically, SD~\cite{StableDiffusion} provides the backbone, where an input image $x\in\mathbb{R}^{H\times W\times 3}$ is encoded into latent $z_0 = E(x)$, perturbed into $z_t$ by Gaussian noise, and denoised across $T$ steps. At each step, a U-Net conditioned on text $p$, pose priors $\varphi$, KB, and DM predicts the noise to reconstruct $\hat{x}$.
\subsection{Knowledge Base}
\label{seckb}
The KB is composed of diverse codebook entries, each encoding distinct visual feature priors. The training of KB consists of two stages: first, aligning image features to the codebook vectors, and second, training a classifier to predict codebook assignments. In the final stage, the trained KB is integrated into the diffusion model to guide image generation.

{\bf{Stage 1: Codebook training.}} The goal of Codebook Training is to learn a discrete representation of image features that can be queried by text. As shown in Fig.~\ref{fig3-a}, similar to VQVAE~\cite{NDRL}, we freeze the CLIP encoder and add a projection layer to process the encoded features, facilitating stable training of the codebook. The codebook consists of trainable entries $\mathcal{Z} = {(e_1, e_2, \dots, e_K)}$, where $\mathcal{Z} \in \mathbb{R}^{K \times C}$, with $K$ denoting the number of codebook entries and $C$ the feature dimension.

Given an image, we encode it to obtain dense image features $V=(v_1,v_2,\cdots,v_N) \in \mathbb{R}^{N \times C}$, where $N = H \times W$ is the number of spatial tokens and $C$ is the feature dimension. Each token $v_i$, $i\in \{1,2,\cdots,N\}$, is quantized via a nearest-neighbor lookup in the codebook embedding space, as defined by the following equation:
\begin{equation}
q(v_i = k \vert V) = 
\begin{cases} 
	1 & \text{if } k = \arg\min_j \left\| v_i - e_j \right\|_2 \\
	0 & \text{otherwise},
\end{cases}
\end{equation}
where $j\in \{1,2,\cdots,K\}$ and $q(v_i = k \vert V)$ denotes a one-hot indicator. We use $q(v_i)$ to represent the index to the corresponding codebook entry. We denote the set of codebook indices for image token as $y=(q(v_1),q(v_2),\cdots,q(v_N))$, which not only allows us to select the quantized image features $V_q=(e_{y_1},e_{y_2},\cdots,e_{y_N})\in \mathbb{R}^{N\times C}$, but also serves as the target labels for subsequent classifier training. Finally, the VQ-VAE~\cite{NDRL} loss consists of a reconstruction term and a commitment term, formulated as:
\begin{equation}
	\label{eq5}
\mathcal{L}_{\text{VQ}} = \left\| \text{sg}[V_q] - V \right\|^2 + \beta \left\| V_q - \text{sg}[V] \right\|^2,
\end{equation}
where sg denotes stop-gradient and $\beta=0.25$. 

After the basic VQ-VAE objective, we encourage balanced codebook usage. Without regularization, training often collapses to a few entries. To mitigate this, we add an entropy loss that promotes diverse token assignments. Formally, recall that the quantization process yields the index set of image tokens $y=(q(v_1),q(v_2),\cdots,q(v_N))$, where each $q(v_i)$ corresponds to the selected codebook entry. To measure the usage distribution of the codebook, we define the empirical frequency of token assignments as:
\begin{equation}
	p_k = \frac{1}{N} \sum_{i=1}^N \mathbf{1}[q(v_i) = k], \quad p \in \mathbb{R}^K,
\end{equation}
where $K$ denotes the number of codebook entries and $p_k$ is the average usage probability of the $k$-th entry. Based on this distribution, we compute the Shannon entropy:
\begin{equation}
	\mathcal{H}(p) = -\sum_{k=1}^K p_k \log (p_k + \epsilon),
\end{equation}
where $\epsilon$ is a small constant for numerical stability. The entropy is then normalized by the maximum entropy $\log K$, yielding a value in $[0,1]$:
\begin{equation}
	\mathcal{H}_{\text{norm}}(p) = \frac{\mathcal{H}(p)}{\log K}.
\end{equation}	
Finally, the entropy regularization loss is defined as:
\begin{equation}
	\label{eq:entropy}
	\mathcal{L}_{\text{entropy}} = 1 - \mathcal{H}_{\text{norm}}(p),
\end{equation}
which penalizes skewed token usage and encourages uniform allocation across all codebook entries. The overall codebook training objective becomes:
\begin{equation}
	\mathcal{L} = \mathcal{L}_{\text{VQ}} + \mathcal{L}_{\text{entropy}}.
\end{equation}

{\bf{Stage 2: Classifier training.}} This stage trains a classifier to align text features with the learned discrete visual tokens (codebook entries). As shown in Fig.~\ref{fig3-b}, Given an input text, we extract a embedding feature $F$ using a freezed pretrained text encoder (e.g., CLIP). Next is the setting of the {\bf classifier}. The dimension of the result after $F$ is flattened and projected is changed:
\begin{equation}
	X = \text{Linear}(\text{flatten}(F)).
\end{equation}
where $X\in \mathbb{R}^{N\times V}$. $N$ matches the number of image tokens, hence the same symbol and $V$ is feature dimension. subsequently, like the operation of PCTPose~\cite{PCTPose}, four MLP-Mixer~\cite{Mlp_mixer} blocks is used to process the features $X$, and output the logits of token classification:
\begin{equation}
	\hat{y} = \mathcal{M}(X),
\end{equation}
where $\hat{y}$ has the shape of $\mathbb{R}^{N\times K}$ and $K$ is the number of ocdebook entries. The supervision $y$ is obtained from a pretrained and frozen stage 1, which takes the input image corresponding to the current text inputs. We optimize $\hat{y}$ against $y$ using cross-entropy loss:
\begin{equation}
	\label{eq7}
	\mathcal{L}_\text{cls} = \text{CE}(y, \hat{y}).
\end{equation}
 This learning process encourages the alignment of textual features with the corresponding visual codebook entries, thus enabling text-conditioned visual token retrieval. When this stage is trained, the inference stage is shown by the yellow arrow of Fig.~\ref{fig3-b}, where $V_q$ is the retrieved result.
\begin{table*}[!t]
	\centering
	\caption{Results on Human-Art dataset. Methods with * are evaluated on released checkpoints.}
	
	%		$\dagger$ indicates that the LAION-Human dataset~\cite{Humansd} used in our experiments is a self-selected subset,	and ${}^{\text{R}}$ denotes results obtained by replicating their code on our chosen version of LAION-Human.}
\label{tb1}
\begin{tabular}{l l c c c c c c}
	\toprule
	\multirow{2}{*}{Dataset} & \multirow{2}{*}{Method} & \multicolumn{3}{c}{Pose Accuracy} & \multicolumn{2}{c}{Image Quality} & \multicolumn{1}{c}{T2I Alignment} \\
	\cmidrule(lr){3-5} \cmidrule(lr){6-7} \cmidrule(lr){8-8}
	& & AP(\%) $\uparrow$ & CAP(\%) $\uparrow$ & PCE $\downarrow$ & FID$\downarrow$ & KID $\downarrow$ & CLIP-score(\%) $\uparrow$ \\
	\midrule
	\multirow{9}{*}{Human-Art} 
	& SD*~\cite{StableDiffusion} & 0.24 & 55.71 & 2.30 & 11.53 & 3.36 & 33.33 \\
	& T2I-Adapter~\cite{T2iAdapte} & 27.22 & 65.65 & 1.75 & 11.92 & 2.73 & 33.27 \\
	& ControlNet~\cite{ControlNet} & 39.52 & 69.19 & 1.54 & 11.01 & 2.23 & 32.65 \\
	& Uni-ControlNet~\cite{UniControlnet} & 41.94 & 69.32 & 1.48 & 14.63 & 2.30 & 32.51 \\
	& GLIGEN~\cite{Gligen} & 18.24 & 69.15 & 1.46 & -- & -- & 32.52 \\
	& HumanSD~\cite{Humansd} & 44.57 & 69.68 & 1.37 & 10.03 & 2.70 & 32.24 \\
	& GRPose~\cite{Grpose}& 49.50 & 70.84& 1.43 & 13.76 & 2.53 & 32.31 \\
	& Stable-Pose~\cite{StablePose}& 48.88 & 70.83 & 1.50 & 11.12 & 2.35 & 32.60 \\
	\cmidrule(lr){2-8}
	&KB-DMGen &\textbf{53.47}  & \textbf{72.33} & 1.56& 10.54 &2.54&32.43\\
	
	%		    \midrule
	%		    \multirow{3}{*}{LAION-Human$^\dagger$} 
	%		& GRPose$^\text{R}$ & && \\
	%			&KB-DMGen* &&&\\
	\bottomrule
\end{tabular}
\end{table*}
{\bf{Stage3: KB Embedding Diffusion Model.}} In this stage, the pretrained KB is integrated into a diffusion model to guide the image generation process. Fig.~\ref{fig2} illustrates that the text $p$ is encoded by text encoder $\tau_\theta$ to query the KB and generate visual semantic priors $z_t^{''}$. During stage-2 inference (Fig.\ref{fig3-b}), the feature $V_q$ is equivalent to $z_t^{''}$. To achieve the integration of $z_t^{''}$ and U-Net, we employ a FiLM-style~\cite{FiLM_style} modulation block. Concretely, the feature $z_t^{''}$ retrieved from the codebook $\mathcal{Z}$ is passed through an Multi-Layer Perceptron (MLP)~\cite{MLP,DeepLearning} to generate the affine parameters $(\gamma_{cb}, \beta_{cb})$. Meanwhile, the diffusion time step $t$ embedding $\tau_t\in \mathbb{R}^{C_t}$ is projected to generate $(\gamma_t, \beta_t)$, which serves as a global modulation to further ensure overall image quality. The final modulation parameters are obtained by combining both sources:
\begin{equation}
	\label{eqgammtbetat}
	\gamma = \gamma_{cb} \odot \gamma_t, \quad \beta = \beta_{cb} \odot \beta_t,
\end{equation}
where $\odot$ means element-wise multiplication. The $z_t^{'''}$ in Fig.~\ref{fig2} can be obtained and injected U-Net:
\begin{align}
		\label{gammbeta}
	z_t^{'''}=z_t \odot (1 + \gamma) + \beta,
\end{align}
where $(\gamma,\beta)=\text{MLP}(z_t^{''},\tau_t)$ are adapted by both visual priors $z_t^{''}$ and time step $t$. Therefore, the KB-conditioned image generation $\kappa_\theta$ can be formulated as:
\begin{align}
	\kappa_\theta(z_t, \mathcal{Z},t)& = z_t \odot (1 + \gamma) + \beta ,
\end{align}
where $\mathcal{Z}$ denotes the codebook set in KB. 

\subsection{Dynamic Masking}
\label{secdm}
Our method builds upon the coarse-to-fine Pose-Masked Self-Attention (PMSA) of Stable-Pose~\cite{StablePose}, which applies Gaussian-dilated pose masks to gradually refine latent representations. The difference is that we combine Dynamic Mask and PMSA (DM-PMSA).

{\bf DM-PMSA.} A binary pose mask $m_k$ is obtained from the skeleton image, downsampled to match the latent feature $z_t\in \mathbb{R}^{c\times h\times w}$ (see Fig.~\ref{fig2}), and dilated by Gaussian kernels of decreasing sizes $\{k_1 > \cdots > k_N\}$---we directly adopt the optimal Gaussian kernel configuration $\{23,13\}$ with standard deviation $\sigma=3$ from~\cite{StablePose}---to produce ${m_{k_1}, \dots, m_{k_N}}$. The latent feature $z_t$ is then processed by a sequence of $N=2$ ViT blocks, each associated with one of the Gaussian-dilated masks in a coarse-to-fine manner. Within each block, the all patch embeddings of $z_t$ are projected into querie $Q$, key $K$, and value $V$, and standard attention logits~\cite{AttentionIsAllYouNeed} are computed as:
\begin{equation}
	\text{dots} = \frac{QK^\top}{\sqrt{d}},
\end{equation}
where $d$ is the projected channel. Then, Attention is restricted to pose-relevant regions and dynamically modulated:
\begin{equation}
	\label{eqdm}
	A_k = \text{softmax}\left\{(1+\delta m_k)(\text{dots} + \text{AttnMask}(m_k)\right)\}V,
\end{equation}
where $\delta = \text{Sigmoid}\{\textbf{MLP}(t)\}$ is a timestep-dependent modulation factor applied only to pose regions via $\delta m_k$, and $\text{AttnMask}(m_k) \in \mathbb{R}^{l \times l}$ where $l=h\times w$ is a pose-aware attention mask: entries corresponding to pose-related patches ($m_k=1$) are set to $0$, while all other entries are assigned $-\infty$, suppressing attention outside pose regions. This coarse-to-fine progression of $N$ blocks gradually steers the latent representation to align with the target pose, while the dynamic modulation provides additional flexibility in controlling pose influence across timesteps. 

We define $ F_\theta^\text{dyn}$ as the DM-PMSA process, and the {\bf conditioning function} $\nu_\theta(z_t, \varphi, t)$ as the combination of the DM-PMSA and the pose encoder:
\begin{align}
	\nu_\theta(z_t, \varphi, t) &= F_\theta^\text{dyn}(z_t, \varphi, t) + \beta_\theta(\varphi),\\
z_t^{'}&=z_t+	\nu_\theta(z_t, \varphi, t)
\end{align}
where $\varphi\in\mathbb{R}^{h\times w\times 3}$ is  the input pose skeleton in Fig.~\ref{fig2}. In this way, $	\nu_\theta$ captures both the spatial dependencies between body parts (via $F_\theta^\text{dyn}$ ) and global pose features (via $\beta_\theta$), with adaptive modulation across diffusion timesteps. Following Stable-Pose, $\beta_\theta$  is a trainable encoder, including six convolutional layers with SiLU activation layers,  downsampling the input pose image by a factor of 8. A zero-convolutional layer is added in the end.

As shown in Fig.~\ref{fig2}, in our framework, the injection of KB and DM is not performed at the same stage. Specifically, the DM is introduced at the early stage of the first layer ($L_{inj}=\{1\}$), leveraging the rich early image features to provide fine-grained control over local pose and structural information. In contrast, the KB is injected at intermediate and later layers ($L_{inj}=\{4,7,10\}$), providing global visual knowledge guidance in the mid-to-late stages.

\subsection{Loss of KB-DMGen}
\label{loss}
As shown in Fig.~\ref{fig2},  the denoising network $\epsilon_\theta$ adopts a UNet backbone. Let $\epsilon_\theta(\mathbf{z}_t, t, p),\ t \in \{1, \cdots, T\}$, represent a T-step denoising UNet with gradients $\nabla\theta$ over a batch and input text prompt $p$. The denoising model predicts the noise as:
\begin{equation}
	\epsilon_\text{pred} = \epsilon_\theta(z_t, t, \tau_\theta (p), \nu_\theta(z_t, \varphi,t),\kappa_\theta(z_t,\mathcal{Z},t)),
\end{equation}
where $\tau_\theta$ is the text encoder, $\nu_\theta(z_t, \varphi,t)$ is DM conditioning function in Sec.~\ref{secdm} and $\kappa_\theta(z_t,\mathcal{Z})$ is the KB-guided conditioning function in Sec.~\ref{seckb}.

The reconstruction error outside the pose regions is computed as:
\begin{equation}
	\mathcal{L}_\text{um} = 
\underset{\substack{\mathbf{z}, p, \varphi, \epsilon,\mathcal{Z} \sim \mathcal{N}(0, I), t}}{\mathbb{E}}
	\Big[ \| (\epsilon - \epsilon_\text{pred}) \odot (1 - m_{k_N}) \|_2^2 \Big],
\end{equation}
where $m_{k_N}$ be the finest-level pose mask and $\epsilon$ denote the Gaussian noise added to the latent encoding $z_t$ at timestep $t$. This term ensures that the model maintains consistency in the background and non-pose regions. For pose-relevant areas, the error is measured by:
\begin{equation}
	\mathcal{L}_\text{m} =
	\underset{\substack{\mathbf{z}, p, \varphi, \epsilon, \mathcal{Z} \sim \mathcal{N}(0, I), t}}{\mathbb{E}}
	\Big[ \| (\epsilon - \epsilon_\text{pred}) \odot m_{k_N} \|_2^2 \Big].
\end{equation}
This loss explicitly enforces accurate reconstruction within the pose-constrained regions. Finally, the overall training objective is a weighted sum:
\begin{equation}
	\mathcal{L} = \mathcal{L}_\text{um} + \alpha\mathcal{L}_\text{m},
\end{equation}
where $\alpha = 5$ is a hyperparameter emphasizing the masked (pose) regions, following the optimal design of Stable-Pose~\cite{StablePose}.

\begin{figure*}[t]
	\centering
	
	\includegraphics[width=0.9\linewidth]{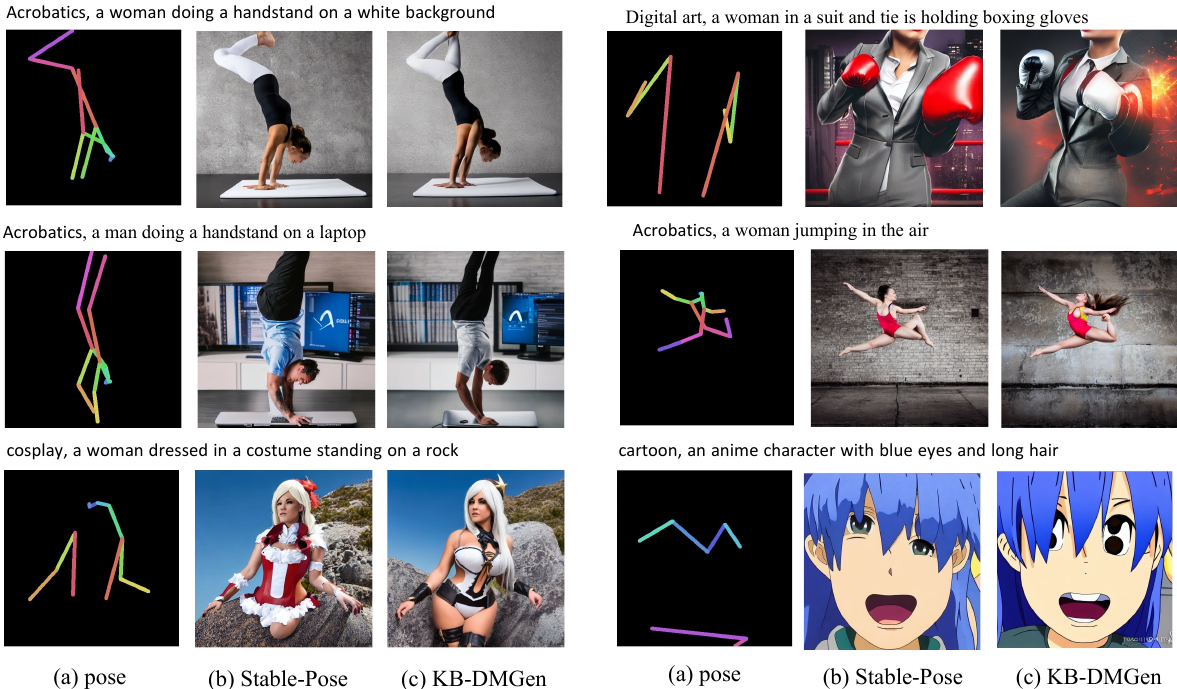}
	
	\caption{The visualization results of  KB-DMGen in single-person scenes.}
	\label{fig4}
\end{figure*}

\begin{figure*}[t]
	\centering
	
	\includegraphics[width=0.9\linewidth]{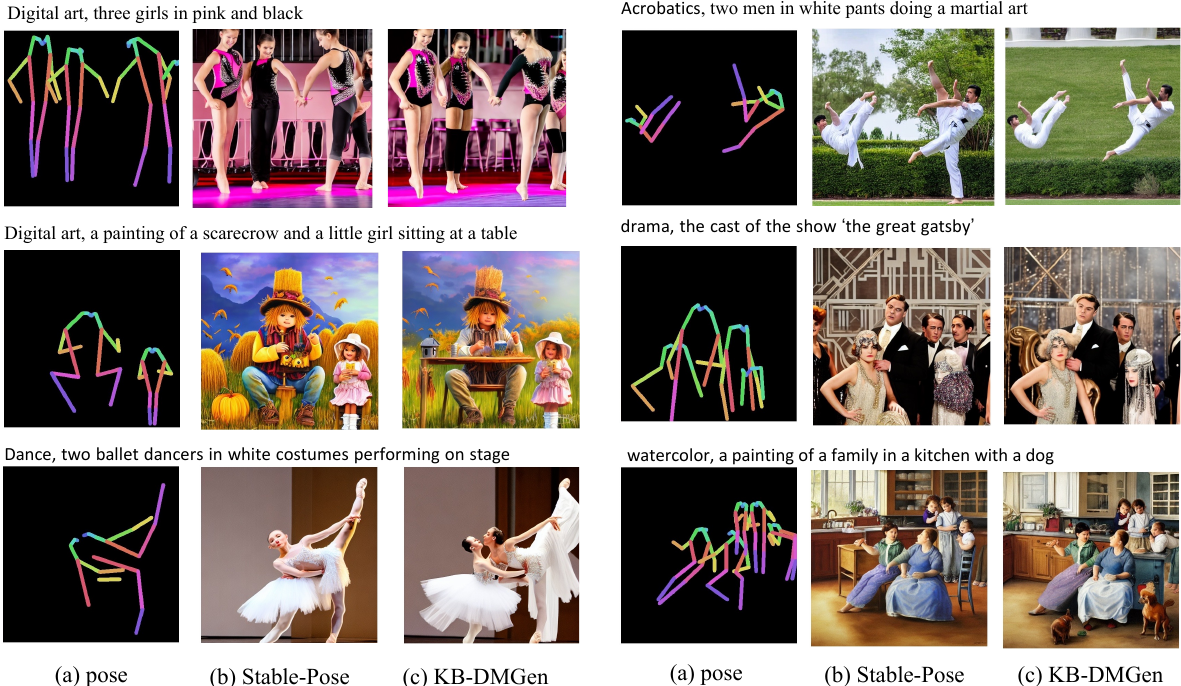}
	
	\caption{The visualization results of  KB-DMGen in multi-person scenes.}
	\label{fig4_1}
\end{figure*}

\begin{figure*}[t]
	\centering
	
	\includegraphics[width=0.9\linewidth]{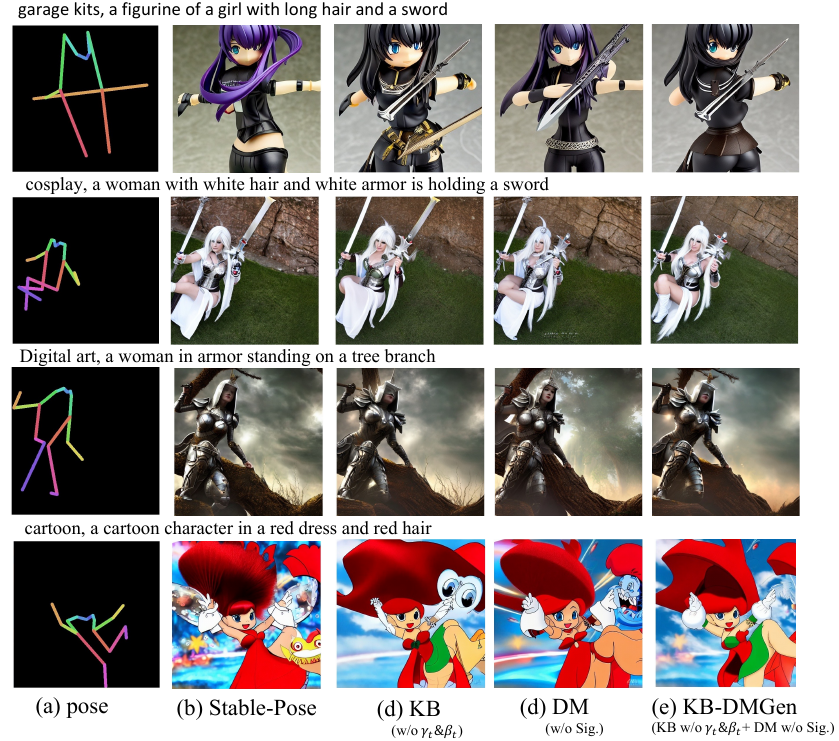}
	
	\caption{Qualitative comparison. The KB (w/o $\gamma_t\&\beta_t$)+DM (w/o Sig.) combination exhibits a synergistic effect, enhancing both pose accuracy and image quality beyond individual contributions. Sig. means Sigmoid.}
	\label{fig4_2}
\end{figure*}

\begin{table*}[!t]
	\centering
	\caption{Ablation study of KB, DM, and their joint effects on Human-Art dataset.}
	\begin{tabular}{l|c|c|c|c|c|c}
		\hline
		Components & AP (\%)$\uparrow$ & CAP (\%)$\uparrow$ & PCE$\downarrow$ & FID$\downarrow$ & KID$\downarrow$ & CLIP-score(\%)$\uparrow$ \\
		\hline
		Stable-Pose~\cite{StablePose} (Base) & 48.88 & 70.83 & 1.50 & 11.12 & 2.35 & 32.60 \\
		+KB & 50.73 & 71.04 & 1.58 & 11.28 & 2.52 & 32.47 \\
		+DM & 49.16 & 70.63 & 1.53 & 11.41 & 2.35 & 32.44 \\
		+KB+DM & 51.40 & 71.17 & 1.54 & 10.56 & 2.54 & 32.41 \\
				+KB+DM+ D\&C &51.71   & 71.40 & 1.53& 10.29 &2.45&32.45\\
		+KB w/o $\gamma_t$+DM w/o Sig. (KB-DMGen)& \textbf{53.47} & \textbf{72.33} & 1.56 & 10.54 & 2.53 & 32.43 \\
		\hline
	\end{tabular}
	\label{tb:main_ablation}
\end{table*}

\begin{table}[!t]
	\centering
	\caption{The KB ablation experiments on the Human-Art dataset.}
	\begin{tabular}{l|c|c|c}
		\hline
		Components & AP (\%)$\uparrow$ & CAP (\%)$\uparrow$ & FID$\downarrow$  \\
		\hline
		Base & 48.88 & 70.83  & 11.12   \\
		+KB & 50.73 & 71.04  & 11.28 \\
		+KB(w/o $\mathcal{L}_{\text{entropy}}$) & 50.61 & 71.16  & 11.33  \\
		+KB(w/o $\gamma_{cb}\&\beta_{cb}$) & 48.34 & 70.57 & 11.70 \\
		+KB(w/o $\gamma_t\&\beta_t$) &\textbf{51.12} & \textbf{71.70} & 11.85  \\
		\hline
	\end{tabular}
	\label{tb:kb_ablation}
\end{table}
\begin{table}[!t]
	\centering
	\caption{DM ablation experiments on Human-Art dataset.}
	\begin{tabular}{l|c|c|c}
		\hline
		Components & AP (\%)$\uparrow$ & CAP (\%)$\uparrow$ & FID$\downarrow$\\
		\hline
		Base & 48.88 & 70.83 & 11.12  \\
		+DM & 49.16 & 70.63  & 11.41 \\
		+DM(w/o sigmoid) &  \textbf{49.36} & \textbf{71.43} & 12.45\\
		\hline
	\end{tabular}
	\label{tb:dm_ablation}
\end{table}

\begin{table*}[!t]
	\centering
	\caption{The number of codebook entries $K$ Ablation experiments on Human-Art dataset}
	\begin{tabular}{l|c|c|c|c|c|c}
        \hline  % 代替 \toprule
		$K$ & AP (\%)$\uparrow$ & CAP (\%)$\uparrow$ & PCE$\downarrow$ &FID$\downarrow$ &KID$\downarrow$&CLIP-score(\%)$\uparrow$\\
        \hline  % 代替 \midrule
		256&50.09&71.06&1.56&11.34&2.47&32.48\\
		512 & 50.35& 70.85&1.56&11.11&2.44&32.45\\
		1024 & {\bf 50.73}& 71.04 & 1.58&11.28&2.52&32.47 \\
		2048&50.40&71.08&1.52&11.49&2.53&32.47\\
		
		        \hline  % 代替 \midrule
		
	\end{tabular}
	\label{tb:codebook_size}
\end{table*}

\begin{table}[t]
	\centering
	\caption{Number of trainable parameters across different stages and the number of codebook entries $K$ in KB.}
	\begin{tabular}{c|cccc}
		\hline
		\multirow{2}{*}{Stage} & \multicolumn{4}{c}{Trainable Parameters (M)} \\
		\cline{2-5}
		&  $K=256$&$K=512$ & $K=1024$ & $K=2048$\\
		\hline
		 1 & 9.70 & 9.96&10.49&11.54\\
		 2 & 78.33 & 78.35& 78.38&78.45\\
		\hline
	\end{tabular}
	\label{tab7}
\end{table}

\begin{table}[t]
	\centering
	\caption{Number of parameters of FiLM across different inject layers $L_{inj}$ in KB-DMGen.}
	\begin{tabular}{c|cccc}
		\hline
		$L_{inj}$ & $\{4\}$ & $\{7\}$ & $\{10\}$ & all$\{4,7,10\}$ \\
		\hline
		{Params (M)} & 10.8& 28.2 & 28.2& 67.3\\
		\hline
	\end{tabular}
	\label{tab8}
\end{table}

\section{Experiments}
\subsection{Experimental Settings}
{\bf{Datasets}}
%We evaluated our model on the Human-Art~\cite{HumanArt} and LAION-Human~\cite{Humansd} datasets. 
We evaluated our model on the Human-Art~\cite{HumanArt}. The Human-Art dataset comprises 50,000 high-quality images from 5 real-world and 15 virtual scenarios, featuring human bounding boxes, key points and textual descriptions. The LAION-Human dataset~\cite{Humansd} consists of approximately
1 million image-caption pairs, filtered by high image quality and human estimation confidence scores, using a diverse range of human activities and more realistic images.

% In the LAION-Human dataset, we randomly selected 200,000 samples for training and 20,000 for testing. Since the original splits of GRPose~\cite{Grpose} and StablePose~\cite{StablePose} are unknown and some images fail to download, our samples differ, but the quantities remain the same, and we re-implemented both methods and trained and evaluated them on our split.

{\bf{Implementation Details}}
Similar to previous work~\cite{ControlNet,T2iAdapte,StablePose}, we fine-tune our model on SD with version 1.5. We utilize Adam~\cite{Adam} optimizer with a learning rate of $1\times10^{−5}$. We also follow~\cite{ControlNet} to randomly replace text prompts as empty strings at a probability of 0.5, which aims to strengthen the control of the pose input. During inference, no text prompts are removed and a DDIM sampler~\cite{DDIM} with 50 time steps is utilized to generate images. We train our models with 10 epochs on all datasets. The number of codebook entried $K$ is set 1024.  The batch size of inference is same as Stable-Pose~\cite{StablePose}. 
%For the Human-Art dataset~\cite{HumanArt}, the training is executed using five NVIDIA A6000 GPUs with 1 batch size. For our filtered LAION-Human dataset, training is conducted using five NVIDIA A6000 GPUs with a batch size of 8. Total training durations are 50,000 hours for our reproduction of StablePose~\cite{StablePose}, 70,000 hours for our KB-DMGen, and 50,000 hours for our reproduction of GRPose~\cite{Grpose}.
%下图对应的输入：cartoon,two cartoon characters on top of a roof
\begin{figure*}[htbp]
%			\label{Linj4}
		\subfloat[$L_{inj}=\{4\}$]{
				\begin{tikzpicture}[scale=0.65]
				% 主坐标轴：左Y轴和底部X轴
				\begin{axis}[
					x dir=reverse,
					xlabel={Sampling Step},
					xlabel style={font=\small},
					ylabel={$\gamma\times10^2$},
					ylabel style={font=\small,anchor=east, at={(0.05,0.65)}},
					axis y line*=left,
					axis x line*=bottom,
					legend style={font=\scriptsize,at={(0.1,0.9999)},anchor=north west,draw=black!20,
						draw=none,        % 无边框
						fill=none,       % 透明背景
						inner sep=1pt,    % 最小边距
					},
					legend cell align={left},
					]
					% 参数数量（左轴）
					\addplot[red,mark=*,mark size=1.25pt,solid,mark repeat=4] table[x=timestep, y expr=\thisrow{gamma}*100, col sep=comma]{32.csv};
%					\addlegendentry{ $\gamma$}
					
					 \addplot[blue,mark=triangle*,mark size=1.25pt,solid,mark repeat=4] 
					table[x=timestep, y expr=\thisrow{gamma}*100, col sep=comma]{32_KB.csv};
%					\addlegendentry{ $\gamma$}
				\end{axis}
				
				% 第二坐标轴：右Y轴和顶部X轴
				\begin{axis}[
					ylabel={$\beta\times10^2$},
					ylabel style={font=\small,anchor=west, at={(1.25,0.3)}},
					axis y line*=right,
					axis x line*=top,
					xtick=\empty, % 避免重复 xtick
					legend style={font=\scriptsize,at={(0.1,0.9554)},anchor=north west,draw=black!20,
						draw=none,        % 无边框
						fill=none,       % 透明背景
						inner sep=1pt,    % 最小边距
					}
					]
					% 计算量（右轴）
					\addplot[red,mark=*,mark size=1.25pt, dashed,mark repeat=4] table[x=timestep, y expr=\thisrow{beta}*100, col sep=comma]{32.csv};

					\addplot[blue,mark=*,mark size=1.25pt, dashed,mark repeat=4] table[x=timestep, y expr=\thisrow{beta}*100, col sep=comma]{32_KB.csv};
				
%					\addlegendentry{$\beta$}
				\end{axis}
\node[anchor=north west] at (2,5.9) { % 根据需要调整位置
	\begin{tikzpicture}[scale=0.4]
		\tiny
		% 设置行间距
		\def\dy{0.54} % 每行的垂直间距
		% 红色实线 gamma
		\draw[red,solid,line width=0.4pt] (0,0) -- (0.35,0);
		\node[right] at (0.38,0) {$\gamma$ (KB+DM)};
		% 红色虚线 beta
		\draw[red,line width=0.4pt, dash pattern=on 1pt off 1pt] (0,-\dy) -- (0.35,-\dy);
		\node[right] at (0.38,-\dy) {$\beta$ (KB+DM)};
		% 蓝色实线 gamma
		\draw[blue,solid,line width=0.4pt] (0,-2*\dy) -- (0.35,-2*\dy);
		\node[right] at (0.38,-2*\dy) {$\gamma$ (KB)};
		% 蓝色虚线 beta
		\draw[blue,line width=0.4pt, dash pattern=on 1pt off 1pt] (0,-3*\dy) -- (0.35,-3*\dy);
		\node[right] at (0.38,-3*\dy) {$\beta$ (KB)};
	\end{tikzpicture}
};
				\end{tikzpicture}}
%			\label{Linj7}
			\subfloat[$L_{inj}=\{7\}$]{
					\begin{tikzpicture}[scale=0.65]
					% 主坐标轴：左Y轴和底部X轴
					\begin{axis}[
						x dir=reverse,	
						xlabel={Sampling Step},
						xlabel style={font=\small},
						ylabel={$\gamma\times10^2$},
						ylabel style={font=\small,anchor=east, at={(0.085,0.65)}},
						axis y line*=left,
						axis x line*=bottom,
						legend style={font=\scriptsize,at={(0.1,0.9999)},anchor=north west,draw=black!20,
							draw=none,        % 无边框
							fill=none,       % 透明背景
							inner sep=1pt,    % 最小边距
						},
						legend cell align={left},
						]
			
						\addplot[red,mark=*,mark size=1.25pt, solid,mark repeat=4]table[x=timestep, y expr=\thisrow{gamma}*100, col sep=comma]{16.csv};
						
							\addplot[blue,mark=*,mark size=1.25pt, solid,mark repeat=4]table[x=timestep, y expr=\thisrow{gamma}*100, col sep=comma]{16_KB.csv};
		
%						\addlegendentry{ $\gamma$}
					\end{axis}
					
					% 第二坐标轴：右Y轴和顶部X轴
					\begin{axis}[
						ylabel={$\beta\times10^2$},
						ylabel style={font=\small,anchor=west, at={(1.25,0.28)}},
						axis y line*=right,
						axis x line*=top,
						xtick=\empty, % 避免重复 xtick
						legend style={font=\scriptsize,at={(0.1,0.9554)},anchor=north west,draw=black!20,
							draw=none,        % 无边框
							fill=none,       % 透明背景
							inner sep=1pt,    % 最小边距
						}
						]
	
						\addplot[red,mark=*,mark size=1.25pt, dashed,mark repeat=4] table[x=timestep, y expr=\thisrow{beta}*100, col sep=comma]{16.csv};
						
						\addplot[blue,mark=*,mark size=1.25pt, dashed,mark repeat=4] table[x=timestep, y expr=\thisrow{beta}*100, col sep=comma]{16_KB.csv};
					
%						\addlegendentry{$\beta$}
					\end{axis}
					\node[anchor=north west] at (2,5.9) { % 根据需要调整位置
						\begin{tikzpicture}[scale=0.4]
							\tiny
							% 设置行间距
							\def\dy{0.54} % 每行的垂直间距
							% 红色实线 gamma
							\draw[red,solid,line width=0.4pt] (0,0) -- (0.35,0);
							\node[right] at (0.38,0) {$\gamma$ (KB+DM)};
							% 红色虚线 beta
							\draw[red,line width=0.4pt, dash pattern=on 1pt off 1pt] (0,-\dy) -- (0.35,-\dy);
							\node[right] at (0.38,-\dy) {$\beta$ (KB+DM)};
							% 蓝色实线 gamma
							\draw[blue,solid,line width=0.4pt] (0,-2*\dy) -- (0.35,-2*\dy);
							\node[right] at (0.38,-2*\dy) {$\gamma$ (KB)};
							% 蓝色虚线 beta
							\draw[blue,line width=0.4pt, dash pattern=on 1pt off 1pt] (0,-3*\dy) -- (0.35,-3*\dy);
							\node[right] at (0.38,-3*\dy) {$\beta$ (KB)};
						\end{tikzpicture}
					};
					
				\end{tikzpicture}}
%			\label{Linj10}
			\subfloat[$L_{inj}=\{10\}$]{
					\begin{tikzpicture}[scale=0.65]
					% 主坐标轴：左Y轴和底部X轴
					\begin{axis}[
						x dir=reverse,
						xlabel={Sampling Step},
						xlabel style={font=\small},
						ylabel={$\gamma\times10^2$},
						ylabel style={font=\small,anchor=east, at={(0.09,0.65)}},
						axis y line*=left,
						axis x line*=bottom,
						legend style={font=\scriptsize,at={(0.1,0.9999)},anchor=north west,draw=black!20,
							draw=none,        % 无边框
							fill=none,       % 透明背景
							inner sep=1pt,    % 最小边距
						},
						legend cell align={left},
						]
			
						\addplot[red,mark=*,mark size=1.25pt, solid,mark repeat=4]table[x=timestep, y expr=\thisrow{gamma}*100, col sep=comma]{8.csv};
						\addplot[blue,mark=*,mark size=1.25pt, solid,mark repeat=4]table[x=timestep, y expr=\thisrow{gamma}*100, col sep=comma]{8_KB.csv};
%						\addlegendentry{ $\gamma$}
					\end{axis}
					
					% 第二坐标轴：右Y轴和顶部X轴
					\begin{axis}[
						ylabel={$\beta\times10^2$},
						ylabel style={font=\small,anchor=west, at={(1.3,0.6)}},
						axis y line*=right,
						axis x line*=top,
						xtick=\empty, % 避免重复 xtick
						legend style={font=\scriptsize,at={(0.1,0.9554)},anchor=north west,draw=black!20,
							draw=none,        % 无边框
							fill=none,       % 透明背景
							inner sep=1pt,    % 最小边距
						}
						]
						\addplot[red,mark=*,mark size=1.25pt, dashed,mark repeat=7] table[x=timestep, y expr=\thisrow{beta}*100, col sep=comma]{8.csv};
							\addplot[blue,mark=*,mark size=1.25pt, dashed,mark repeat=7] table[x=timestep, y expr=\thisrow{beta}*100, col sep=comma]{8_KB.csv};
%						\addlegendentry{$\beta$}
					\end{axis}
						\node[anchor=north west] at (1.1,5.9) { % 根据需要调整位置
						\begin{tikzpicture}[scale=0.4]
							\tiny
							% 设置行间距
							\def\dy{0.54} % 每行的垂直间距
							% 红色实线 gamma
							\draw[red,solid,line width=0.4pt] (0,0) -- (0.35,0);
							\node[right] at (0.38,0) {$\gamma$ (KB+DM)};
							% 红色虚线 beta
							\draw[red,line width=0.4pt, dash pattern=on 1pt off 1pt] (0,-\dy) -- (0.35,-\dy);
							\node[right] at (0.38,-\dy) {$\beta$ (KB+DM)};
							% 蓝色实线 gamma
							\draw[blue,solid,line width=0.4pt] (0,-2*\dy) -- (0.35,-2*\dy);
							\node[right] at (0.38,-2*\dy) {$\gamma$ (KB)};
							% 蓝色虚线 beta
							\draw[blue,line width=0.4pt, dash pattern=on 1pt off 1pt] (0,-3*\dy) -- (0.35,-3*\dy);
							\node[right] at (0.38,-3*\dy) {$\beta$ (KB)};
						\end{tikzpicture}
					};
				\end{tikzpicture}}

			\caption{Visualization of the modulation parameters $\gamma$ and $\beta$ in Eq.~\ref{gammbeta} across the reverse sampling steps ($t=50 \rightarrow 1$) at different injection layers ($L_{\text{inj}}=\{4,7,10\}$). (a) Shallow layers ($L_{inj}=\{4\}$) show strong and rapid variations, mainly guiding coarse structures. (b) Middle layers ($L_{inj}=\{7\}$) exhibit moderate and smoother modulation, transitioning features from coarse to fine structures. (c) Deep layers ($L_{inj}=\{10\}$) remain stable with small amplitude, mainly fine-tuning details. 			\label{figgammabeta}}
\end{figure*}
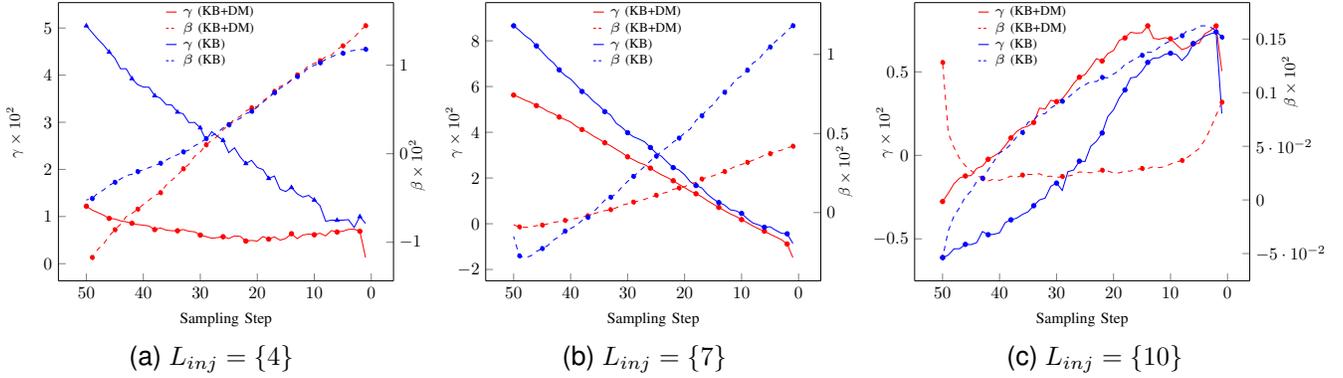

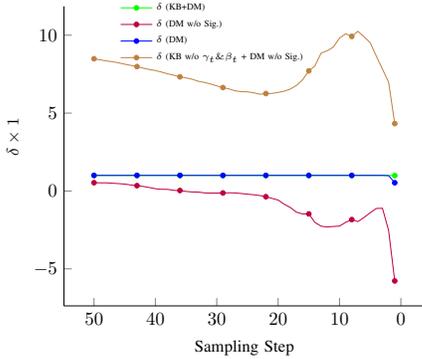
\begin{figure}[t]
	\centering
	\begin{tikzpicture}[scale=0.7]
		% 主坐标轴：左Y轴和底部X轴
		\begin{axis}[
			x dir=reverse,
			xlabel={Sampling Step},
			xlabel style={font=\small},
			ylabel={$\delta\times1$},
			ylabel style={font=\small,anchor=east, at={(0.04,0.65)}},
			axis y line*=left,
			axis x line*=bottom,
			legend style={font=\scriptsize,at={(0.15,1.03)},anchor=north west,draw=black!20,
				draw=none,        % 无边框
				fill=none,       % 透明背景
				inner sep=1pt,    % 最小边距
        row sep=0.001pt,           % 行间距，默认可能偏大，可以调小
        font=\tiny,      % 字体大小，可以改成 \tiny 或				
			},
			legend cell align={left},
			]
			\addplot[green,mark=*,mark size=1.25pt, solid,mark repeat=7]table[x=timestep, y expr=\thisrow{delta}*1, col sep=comma]{mask.csv};
			\addplot[purple,mark=*,mark size=1.25pt, solid,mark repeat=7]table[x=timestep, y expr=\thisrow{delta}*1, col sep=comma]{mask_dmnosig.csv};
			\addplot[blue,mark=*,mark size=1.25pt, solid,mark repeat=7]table[x=timestep, y expr=\thisrow{delta}*1, col sep=comma]{mask_justDM.csv};
			\addplot[brown,mark=*,mark size=1.25pt, solid,mark repeat=7]table[x=timestep, y expr=\thisrow{delta}*1, col sep=comma]{mask_kbnotime_dmnosig.csv};
%			\addlegendentry{ $\delta\times10$}
 \addlegendimage{green,solid,mark=*}
\addlegendentry{$\delta$ (KB+DM)}
\addlegendimage{purple,solid,mark=*}
\addlegendentry{$\delta$ (DM w/o Sig.)}
\addlegendimage{blue,solid,mark=*}
\addlegendentry{$\delta$ (DM)}
\addlegendimage{brown,solid,mark=*}
\addlegendentry{$\delta$ (KB w/o $\gamma_t \& \beta_t$ + DM w/o Sig.)}
		\end{axis}
		
		\end{tikzpicture}
	\caption{Visualization of the DM modulation parameter $\delta$ in Eq.\ref{eqdm} along the reverse sampling steps ($50 \rightarrow 1$).\label{figdelta}}
\end{figure}

For the training of KB, we fine-tune on the OpenCLIP~\cite{OpenClip,OpenClip_software} implementation using the pretrained CLIP model (ViT-L/14) with frozen encoders. Training is performed in two stage\textemdash codebook training and classifier training\textemdash both using the same strategy over 30 epochs with Adam~\cite{Adam} (initial learning rate 0.001), a cosine decay schedule and 256 batch size on per GPU. The training and validation splits as same as Human-Art~\cite{HumanArt}. On the Human-Art dataset, two-stage training takes about 2.5 hours with two NVIDIA 3090 GPUs. 

%On Laion-Human our filtered, it takes about 4 hours with the same GPUs.

{\bf{Metrics.}} To evaluate pose accuracy, we use mean Average Precision (AP), Pose Cosine Similarity-based AP (CAP), and People Counting Error (PCE)~\cite{Kpe}, computed using HigherHRNet~\cite{Higherhrnet} to compare ground-truth poses with those extracted from generated images. For image quality assessment, we employ Fr\'{e}chet Inception Distance (FID)~\cite{FID} and Kernel Inception Distance (KID)~\cite{KID} to measure diversity and fidelity.  KID is multiplied by 100 for Human-Art.Text-image alignment is evaluated using the CLIP-score~\cite{Clip}.

\subsection{Comparison with SOTA Methods}

Quantitative results comparing our method with other state-of-the-art (SOTA) approaches are shown in Table~\ref{tb1}.
 
 {\bf On the Human-Art dataset}, our final model achieves the highest AP and CAP, reaching 53.47 and 72.33 respectively, which surpasses GRPose~\cite{Grpose} by +3.97 AP and +1.49 CAP. Compared with the baseline StablePose~\cite{StablePose}, the improvement is even more substantial (+4.59 AP and +1.50 CAP). Meanwhile, our method maintains competitive global image quality, with FID and KID comparable to other strong baselines. This demonstrates that our design not only significantly improves pose accuracy but also preserves overall image realism. Qualitative results in Fig.~\ref{fig4} and Fig.~\ref{fig4_1} and further confirm that our method produces visually more faithful images with superior pose accuracy and semantic consistency.

%LAION-Human~\cite{Humansd}. We reproduced StablePose and GRPose on a randomly selected subset of the LAION-Human dataset. Our method consistently improves pose alignment and text-image consistency compared to previous approaches.

\subsection{Ablation Studies}
Our method is built upon Stable-Pose~\cite{StablePose}, which serves as the baseline. Ablation studies are conducted on the Human-Art dataset.

{\bf Overall Effects of KB and DM.} We first evaluate the effectiveness of the proposed KB and DM modules, as well as their joint impact on performance. As shown in Table~\ref{tb:main_ablation}, adding KB improves both AP and CAP ensuring image quality. It outperforms the recent SOTA method GRPose~\cite{Grpose} with a 1.23 AP and 0.2 CAP improvement, while achieving better image quality with a 2.48 drop in FID and a slight 0.01 decrease in KID. DM alone mainly enhances AP slightly. Importantly, their combination consistently improves on AP, CAP, FID, while other indicators have only slightly decreased, demonstrating the complementarity between KB and DM. Surprisingly, when removing the temporal scaling factor $\gamma_t\&\beta_t$ from KB in Eq.~\ref{eqgammtbetat} and the sigmoid gating from DM in Eq.~\ref{eqdm}, compared with KB+DM, this method achieves the best AP and CAP with only a slight increase in PCE, demonstrating the complementary role of KB (w/o $\gamma_t\&\beta_t$) global guidance and DM local refinement, as further confirmed by the visualizations in Sec.~\ref{secvis}. What's more, applying component-wise KB querying (+KB+DM+D\&C) achieves the best overall performance compared with KB+DM. Specifically, D\&C leverages the simple and regular structure of HumanArt~\cite{HumanArt} descriptions to decompose each text into three parts---type, object, and status (e.g., "cartoon, an animal character with a sword in the woods" -> type = "cartoon"; object = "an animal character"; status = "with a sword in the woods".). Each component independently queries the KB, and the results are combined into the diffusion model, which enhances semantic parsing by capturing fine-grained correspondences between visual features in KB and textual components.

{\bf Effect of KB.} To further investigate the KB design, we conduct several ablations as reported in Table~\ref{tb:kb_ablation}. Removing the entropy regularization $\mathcal{L}_\text{entropy}$ degrades performance in AP and FID. Eliminating the affine modulation parameters $(\gamma_{cb}, \beta_{cb})$ causes a significant drop in AP and CAP, highlighting their necessity for HIG. Interestingly, removing the timestep-dependent modulation $(\gamma_t, \beta_t)$ yields the highest CAP and AP, but comes at the cost of worse generative fidelity (FID). We omit KID for brevity, as its variance across settings is negligible.

{\bf Effect of DM.} Table~\ref{tb:dm_ablation} summarizes the ablations on the DM module.
While DM improves AP compared to the baseline, removing the sigmoid gating ($\delta = \text{Sigmoid}\{\textbf{MLP}(t)\}$) in Eq.~\ref{eqdm}---where the MLP has only 0.6M parameters---further increases AP and CAP but causes severe degradation of FID by about 1. This indicates that the Sig. (Sigmoid) gate is crucial for stabilizing the modulation strength and preserving global quality, even though it slightly limits precision gains.

{\bf Joint Effects of $\gamma_t\&\beta_t$ and Sigmoid.} The joint analysis in Table~\ref{tb:main_ablation} shows that the combination of KB without $\gamma_t\&\beta_t$ and DM without sigmoid achieves the best AP (53.47) and CAP (72.33). When used individually, each component requires additional constraints ($\gamma_t\&\beta_t$ or sigmoid) to stabilize training and prevent quality degradation, whereas their joint application exhibits complementary benefits. Specifically, KB without $\gamma_t\&\beta_t$ provides robust global guidance, while DM without sigmoid enhances local detail modulation. With global guidance in place, local refinement no longer needs the cautious restriction of sigmoid, and conversely, with flexible local control, global modulation does not require timestep-dependent $\gamma_t\&\beta_t$. Although KB and DM can also exhibit some complementarity, these constraints limit their full potential. Overall, this synergy achieves the best balance between pose precision and image quality.

{\bf Codebook Size.} Finally, we analyze the impact of different codebook sizes $K$ in Table~\ref{tb:codebook_size}.
We observe that the performance first improves as $K$ increases, peaking at $K=1024$, and then slightly decreases with larger codebooks. This indicates that too small a codebook limits the representational capacity, while an excessively large one introduces redundancy and instability. Thus, $K=1024$ provides the best trade-off between efficiency and expressiveness.

\subsection{Visualization of Injection Behaviors}
\label{secvis}
{\bf Visualization of Codebook Injection.} 
In Fig.~\ref{figgammabeta}, we compare the effects of KB and DM on the FiLM parameters $\gamma$ and $\beta$ at different injection layers. KB alone provides strong global trends but shows large fluctuations during sampling. Introducing DM smooths the variations of $\gamma$ and $\beta$, enabling fine-grained local refinement. Different layers exhibit distinct roles: shallow layers use larger modulation to guide coarse structure generation, while deeper layers apply smaller, more stable adjustments to refine details.

{\bf Visualization of Dynamic Mask.} As shown in Fig.~\ref{figdelta}, the curve of DM w/o Sig. gradually transitions from positive to negative during the sampling process, indicating a reversal in the modulation direction of $\delta$. This leads to unstable attention on the pose regions and results in uneven modulation. In contrast, DM effectively enhances and stabilizes the magnitude of $\delta$ in DM and KB+DM. In contrast, guided by the knowledge base, KB (w/o $\gamma_t\&\beta_t$)+DM (w/o Sig.) effectively enhances $\delta$'s magnitude, significantly improving generation accuracy while ensuring overall quality and enabling the dynamic mask to stay focused on key regions even without sigmoid.

{\bf Effect on Generated Results.} 
We further evaluate the qualitative effects of different injection strategies on the generated images (Fig.~\ref{fig4_2}). In this figure, KB (w/o $\gamma_t\&\beta_t$)+DM (w/o Sig.) combination demonstrates a clear synergistic effect, producing more accurate poses and higher-quality images than either component individually. Visualizations of KB and KB (w/o $\gamma_t\&\beta_t$), and DM and DM (w/o Sig.) when used alone, as well as additional KB-DMGen generation results, are provided in the supplementary material.

\subsection{Parameters Analysis}

Table~\ref{tab7} reports the number of trainable parameters when training the knowledge base (KB) alone across its two stages and for different codebook sizes $K$. Increasing $K$ slightly increases parameters in Stage 1, while Stage 2 remains largely unaffected. Table~\ref{tab8} shows the trainable parameters of FiLM modules for different inject layers $L_{inj}$ in KB-DMGen. 
\section{Conclusion}
We propose an image generation method combining a visual KB with DM, validated by experiments. Future work includes expanding the knowledge base and optimizing dynamic masking.

\bibliographystyle{IEEEtran}
\bibliography{bare_jrnl_new_sample4} 
\appendix
\begin{figure*}[!t]
	\centering
	
	\includegraphics[width=0.9\linewidth]{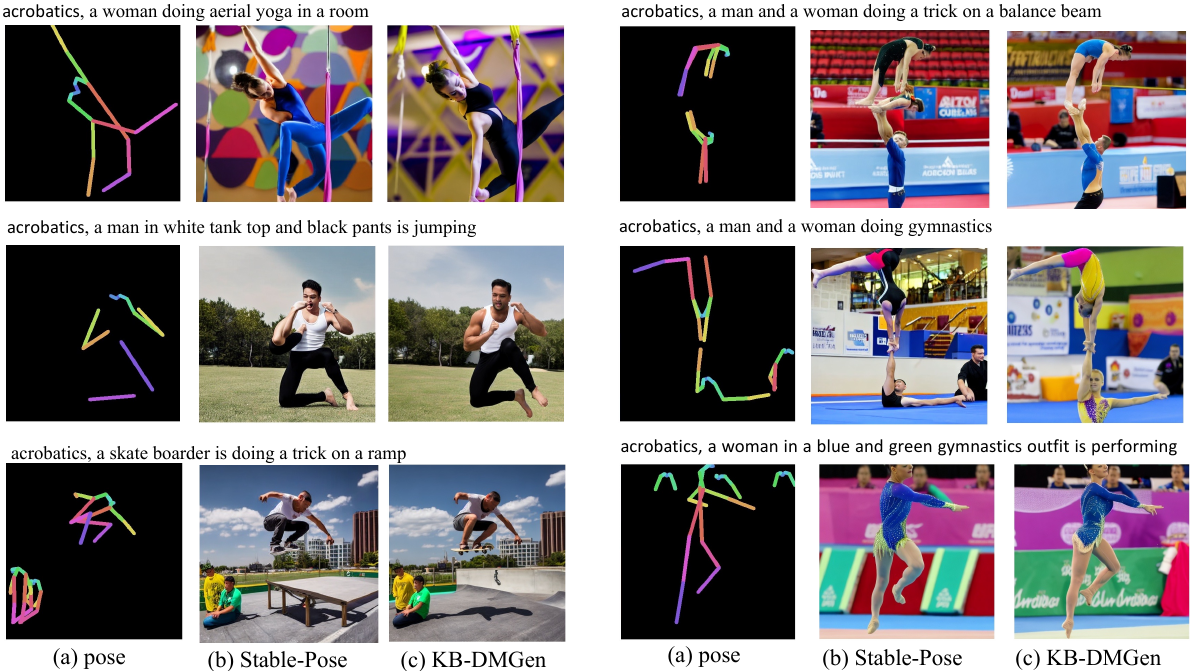}
	
	\caption{Visualization comparison between KB-DMGen and Stable-Pose~\cite{StablePose} on HumanArt datasets~\cite{HumanArt}.}
	\label{figKB_DMGen_extra}
\end{figure*}

\begin{figure*}[!t]
	\centering
	
	\includegraphics[width=0.9\linewidth]{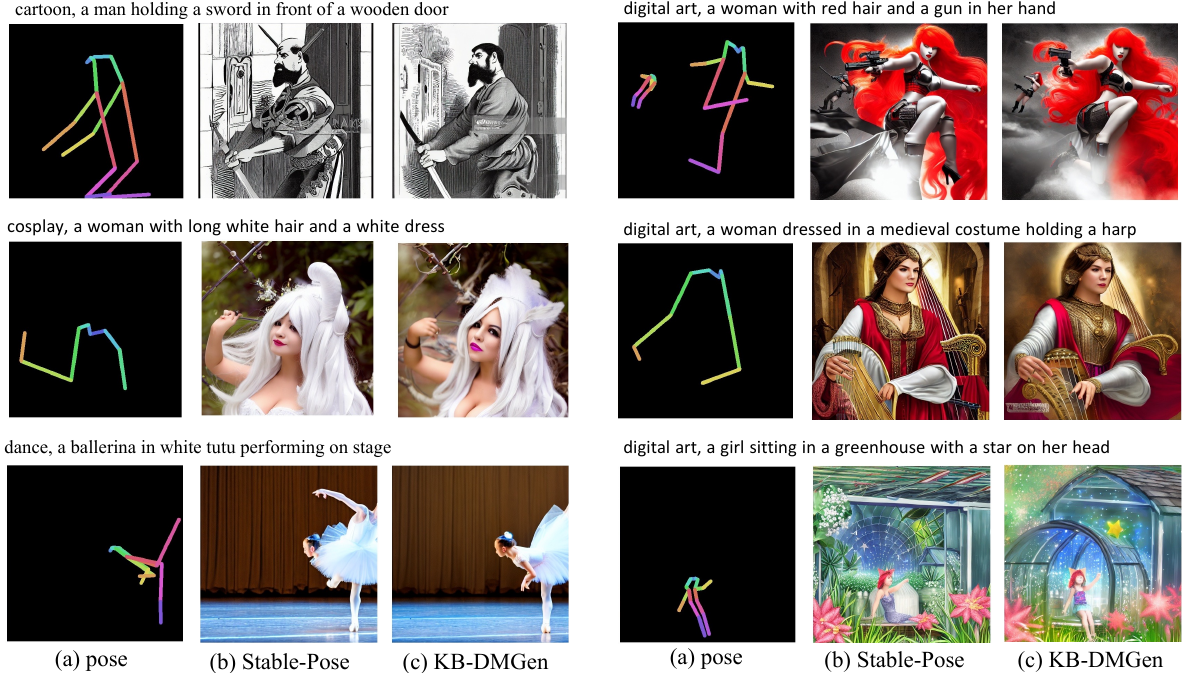}
	
	\caption{Visualization comparison between KB-DMGen and Stable-Pose~\cite{StablePose} on HumanArt datasets~\cite{HumanArt}.}
	\label{figKB_DMGen_extra1}
\end{figure*}
\begin{figure*}[!t]
	\centering
	
	\includegraphics[width=0.9\linewidth]{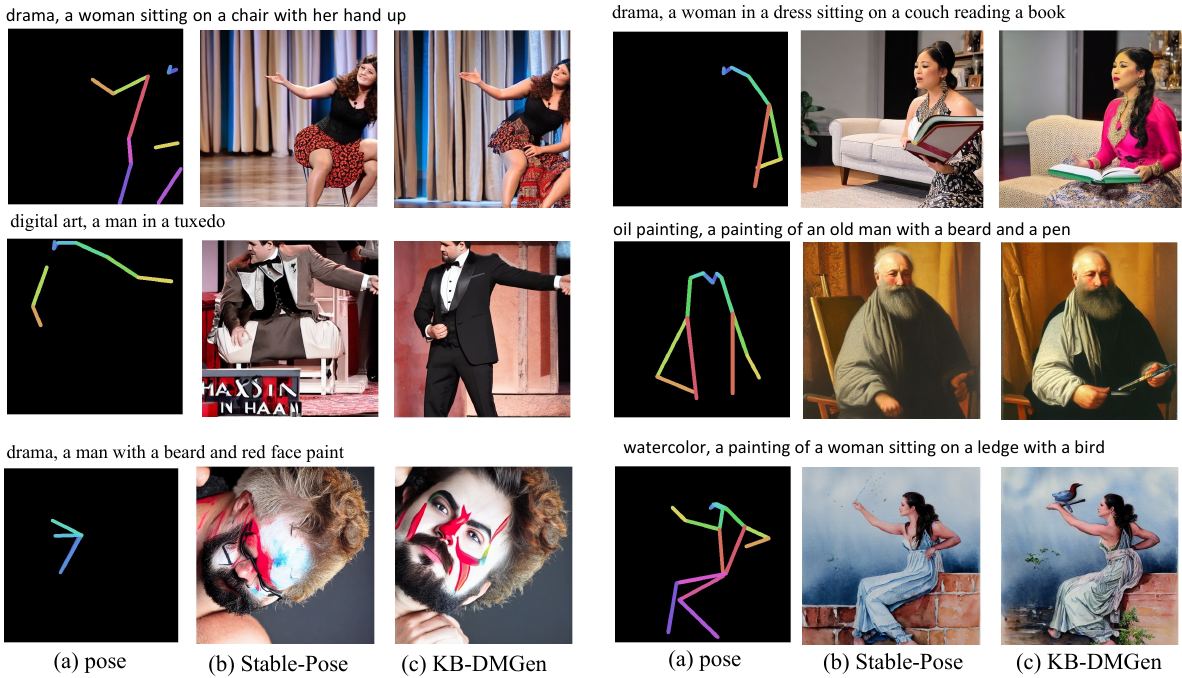}
	
	\caption{Visualization comparison between KB-DMGen and Stable-Pose~\cite{StablePose} on HumanArt datasets~\cite{HumanArt}.}
	\label{figKB_DMGen_extra2}
\end{figure*}

\begin{figure*}[!t]
	\centering
	
	\includegraphics[width=0.9\linewidth]{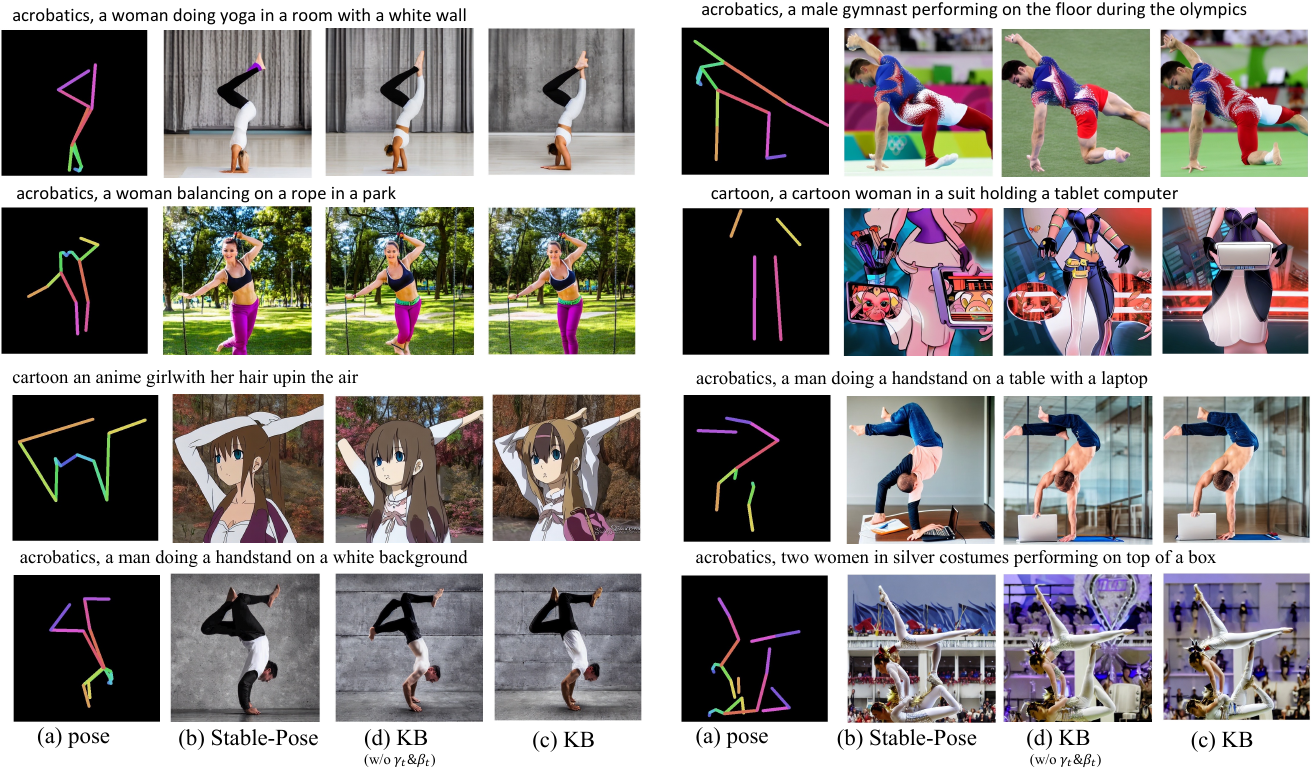}
	
	\caption{Visualization comparison between KB and KB (w/o $\gamma_t\&\beta_t$) on HumanArt datasets~\cite{HumanArt}.}
	\label{figKB_KB_extra}
\end{figure*}
\begin{figure*}[!t]
	\centering
	
	\includegraphics[width=0.9\linewidth]{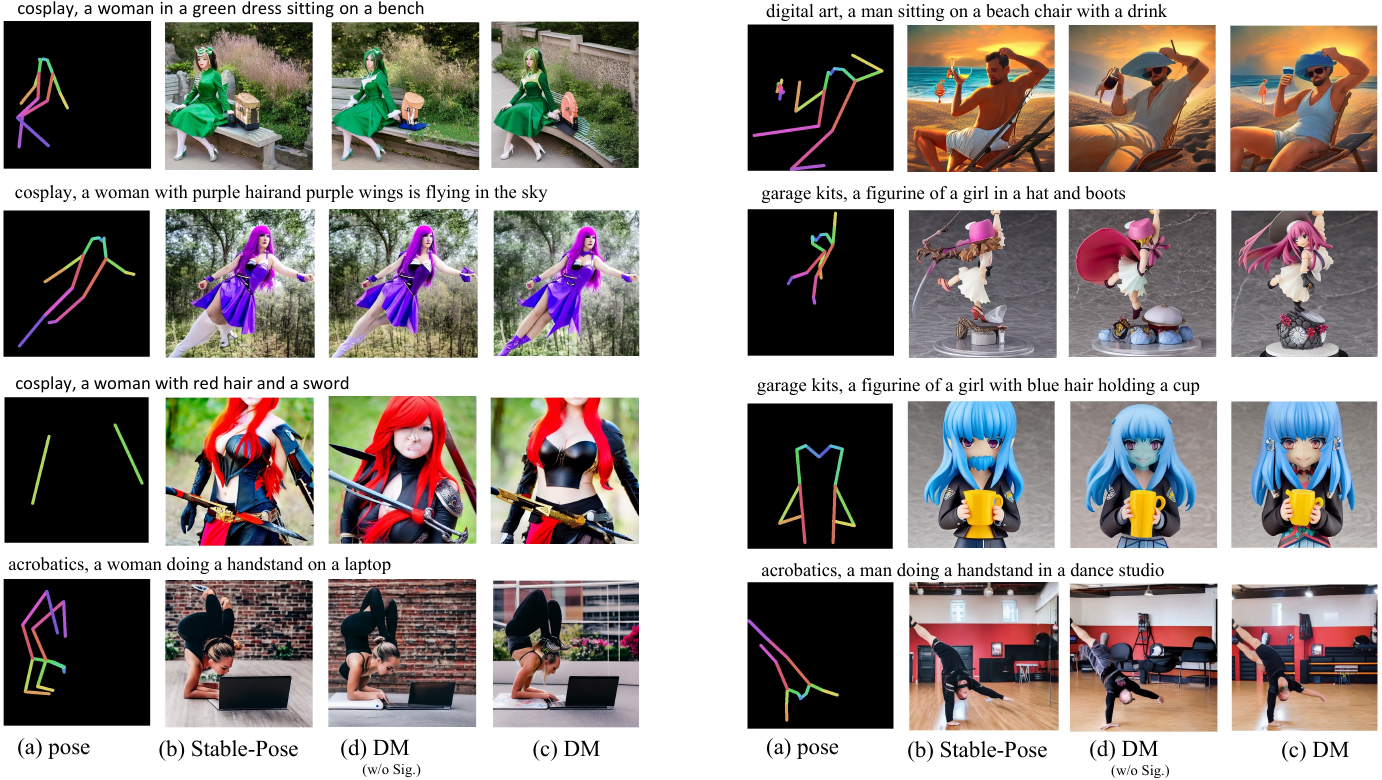}
	
	\caption{Visualization comparison between DM and DM (w/o Sig.) on HumanArt datasets~\cite{HumanArt}.}
	\label{figKB_DM_extra}
\end{figure*}
\section*{KB-DMGen results visualization}
In this section, we provide additional qualitative results for the our method: KB-DMGen. Fig.~\ref{figKB_DMGen_extra}, \ref{figKB_DMGen_extra1} and \ref{figKB_DMGen_extra2} shows a set of example outputs our method achieves improved pose accuracy and overall image quality compared with baselines. 
These additional visualizations further demonstrate the complementary effect of global guidance from KB (w/o $\gamma_t\&\beta_t$) and local refinement from DM (w/o Sig.).

\section*{KB ablation results visualization}
In this section, we focus on visualizing the impact of KB on global quality. Specifically, we compare outputs from KB without affine modulation $(\gamma_t \&\beta_t)$ and standard KB in Fig.~\ref{figKB_KB_extra}. The comparison shows that KB has an advantage in the quality of human image generation, highlighting the importance of the image quality of the temporal modulation pair when using KB alone.

\section*{DM ablation results visualization}
This section visualizes the effect of DM on global quality. 
We compare outputs from DM without sigmoid and standard DM in Fig.~\ref{figKB_DM_extra}. 
The results show that when DM is used alone, the Sigmoid plays an important role in improving the overall quality of the image.

\end{CJK}
\end{document}